\documentclass[review]{elsarticle}

\usepackage{lineno,hyperref}
\modulolinenumbers[5]
\usepackage{mathtools}

\journal{Computer Speech and Language (CSL)}

\begin{document}

\begin{frontmatter}

\title{A Survey of Code-switched Speech and Language Processing}

\author{Sunayana Sitaram}
\address{Microsoft Research India}

\author{Khyathi Raghavi Chandu, Sai Krishna Rallabandi, Alan W Black}
\address{Carnegie Mellon University}




\begin{abstract}
 Code-switching, the alternation of languages within a conversation or utterance, is a common communicative phenomenon that occurs in multilingual communities across the world. This survey reviews computational approaches for code-switched Speech and Natural Language Processing. We motivate why processing code-switched text and speech is essential for building intelligent agents and systems that interact with users in multilingual communities. As code-switching data and resources are scarce, we present a comprehensive list of datasets available in various code-switched language pairs with the language processing tasks they can be used for. We discuss shared tasks and benchmarks that have been proposed to evaluate language processing systems on code-switched text and speech. We review code-switching research in various Speech and NLP applications, including language processing tools and end-to-end systems. We discuss the evaluation of code-switched speech and NLP systems, including recently proposed benchmarks. We conclude with future directions and open problems in the field.
\end{abstract}


\end{frontmatter}

\section{Introduction}



Linguistic code choice refers to the use of a language for a specific communicative purpose and \textbf{code-switching} denotes a shift from one language to another within a single utterance. 
Not only is there a plethora of different languages across the world, but speakers also often mix these languages within the same utterance. 
In fact, some form of code-switching is expected to occur in almost every scenario that involves multilinguals \cite{handbook_languagecontact}. This can go beyond mere insertion of borrowed words, fillers and phrases, and include morphological and grammatical mixing.  Such shifts not only convey group identity \cite{codeswitching_socialidentity}, embody societal patterning \cite{heller1982negotiations} and signal cultural discourse strategies \cite{jacobson2011codeswitching} but also have been shown to reduce the social and interpersonal distance \cite{maltese_distancestudy} in both formal \cite{qian2009codeswitching, rezvani2011code} and informal settings.

In this paper we refer to this phenomenon as {\bf code-switching}, though the term
code-mixing is also used.  While such switching is typically considered informal -
and is more likely to be found in speech and in casual text as now found in social media - it is also found  in semi formal and formal settings such as news paper headlines and teaching. Therefore, we argue that code-switching should not be looked down upon or ignored but be acknowledged as a genuine form of communication that deserves analysis and development of tools and techniques to be handled appropriately. As language technologies improve and permeate more and more applications that involve interactions with humans \cite{peabody2011methods, antol2015vqa}, it is imperative that they take phenomena such as code-switching into account for any consumer facing technology.


Code-switching is most common among peers who have similar fluency in
each language.  For example fluent bilingual Spanish and English
people may often float between their languages, in a form of
communication called Spanglish. Indian sub-continent residents, who
often have a substantial fluency in English will often mix their
speech with their regional languages in Hinglish (Hindi), Tenglish
(Telugu), Tamlish (Tamil) and others. But it is not just English
that code-switching occurs with. Southern Mainland Chinese residents
who, for example, speak Cantonese and Shanghaiese, may switch with
Putonghua (standard Mandarin).  Arabic Dialects are often mixed with
Modern Standard Arabic. The distinction between languages and dialects is hard to define, but we see that code-switching appears with dialects too.  African American Vernacular English (AAVE) speakers will commonly switch between AAVE and Standard American English; Scottish people may switch between Scots and Standard English.  At an extreme, code-switching could also be used to describe register shifting in monolingual speech. Formal speech versus slang or swearing may follow similar functions and patterns as those in code-switching among two distinct languages.

\subsection{Why should we care about code-switched language processing?} 


It is important to realize that humans are good at constructing language registers and learning new
communication methods. Not only are we good at doing this with human-human communication, we also construct and learn to use such registers for human-machine communication efficiently, such as Linux command-line expressions, or the grammar of Alexa interactions.  If we want machines to partake in such human-human conversations, we need to also be able to understand what is being said in these varied registers.  

For the large companies, understanding
code-switched communication will enable better
advertisement-targeting.  Understanding genuine user sentiment about
aspects of products helps improve future versions. \cite{agarwal2017may} found a correlation between language use and sentiment, showing that ignoring one language in favor of the other, or ignoring code-switched languages altogether may lead to the wrong conclusions about user sentiment. For healthcare,
understanding how people feel, if they are being open, will help to
give better care, and enable better communication with patients, and
better distribution and uptake of preventative care.  For educators,
communication in the right register for tutoring, or
understanding if concepts are or are not understood is crucial.  For
entertainment, non-playing characters should communicate in the
appropriate register for the game, and/or be able to understand natural
code-switched communication with other players.

Unlike pidgins or creoles \cite{todd2003pidgins, arends1995pidgins, contact_languages_pidgins_creoles}, where speakers may not have full fluency in the language of influence, we are primarily interested in situations where participants have fluency in each of the languages but are choosing not to stay within one language. Code-switching is not a
simple linguistic phenomena and depending on the languages involved,
and the type of code-switching the interaction between the component
languages may be quite different.  It is easy to identify at least
linguistic sharing, cross-lingual transfer, lexical borrowing as well
as speech errors with restarts commonly within the code-switched data. Likewise although there may be language technology tasks that
can be achieved with straightforward techniques, it is clear that
some tasks, such as semantic role labeling will require complex cross-lingual analysis.


Many have identified the notion of a matrix language in
code-switching \cite{myers2002contact}, that there is an underlying language choice which
mostly defines the grammar and morphological aspects of the utterance.  From a language technologies point of view, especially
when considering code-switched data generation using any form of
language modeling, it is possible to identify `bad' code-switching or
even `wrong' code-switching.  Although it is obviously not a binary
decision, there are extremes that will almost always be wrong.  We
cannot in general randomly choose which language a word would be
realized in, or simply state that we will choose alternate languages
for each word.  That is, there are constraints, there is an underlying grammar and there are multiple linguistic theories that have been proposed for code-switching.  
Modeling the grammar is challenging, even if there may
be an eventual standardized Hinglish that everyone in Northern India may speak, at present, such code-switched languages are very dynamic, and will have very diverse ideolects across speakers.  
This is reminiscent of pidgins and creoles which can develop over time, but they too, especially as they are not normally written languages, are also diverse. 

But we should not give up, there is underlying structure, and there are constraints, and we have good machine learning modeling techniques that can deal with uncertainty. Recently, there has been quite a lot of interest in the speech and NLP community on processing code-switched speech and text, and this paper aims at describing progress made in the field, and discussing open problems. 

This survey is organized as follows. First, we introduce why code-switching is a challenging and important problem for speech and NLP. Next, in Section 2, we briefly describe linguistic studies on code-switching with other theoretical aspects. In Section 3 we describe speech and NLP corpora and resources that have been created for code-switched language pairs. Section 4 describes techniques for building models for code-switching in specific speech and NLP applications. Section 5 describes various shared tasks and challenges that have been conducted to evaluate code-switching, and introduces benchmarks that evaluate models across tasks and languages. We conclude in Section 6 with a description of the challenges that remain to be addressed and future directions. 

\section{Background}

Research on code-switching is not recent, and this phenomenon has been studied by linguists for decades. In this section, we provide a description of linguistic studies on code-switching and how to characterize code-switched languages. We do not attempt to be comprehensive, since providing a complete description of code-switching research in linguistics is out of the scope of this paper.

\subsection{Types of code-switching}

Code-switching is defined as the `juxtaposition within the same speech exchange of passages of speech belonging to two different grammatical systems or sub-systems'\cite{gumperz1982discourse}, while code-mixing is `the embedding of linguistic units such as phrases, words and morphemes of one language into an utterance of another language'\cite{myers1997duelling}. The distinction between code-switching, mixing and lexical borrowing is often not clear and can be thought of as lying on a continuum \cite{bali2014borrowing}. In this paper, we use the terms `code-switching' and `code-mixing' interchangeably, although the distinction between the two may be important for certain applications.

The extent and type of code-switching can vary across language pairs. \cite{rijhwani2017estimating} used word-level Language Identification to estimate which language pairs were code-switched on Twitter. They found that around 3.5\% of tweets were code-switched, with the most common pairs being English-Spanish, English-French and English-Portuguese. English-German tweets typically had only one switch point, implying that the tweets usually contained translations of the same content in English and German, while English-Turkish tweets had the most switch points, implying fluid switching between the two languages. Code-switching can also vary within a language pair. For example, casual conversational Hinglish may be different from Hinglish used in Bollywood movies, which may be different from Hinglish seen on Twitter. 

\subsection{Linguistic Models of Code-switching}

Early approaches investigate code-switching by laying down a formal framework taking into account the two grammatical systems of the languages being mixed and a mechanism to switch between these two systems at the intra-sentential level \cite{joshi1982processing}. This model mainly explores asymmetric relations between the two grammars, without an explicit formalism of a third grammar and the understanding of where and how to switch closed class items.

Quantitative analysis conducted by \cite{poplack1978syntactic} revealed two constraints (1) Free Morpheme Constraint and (2) Equivalence Constraint that function simultaneously. The Free Morpheme constraint specifies that it is possible to switch between full sentences as well as any constituent within the sentence if a free morpheme is present in a constituent. The Equivalence Constraint specifies that language switches  generally occur at points where there is no violation of syntactic rules of the participating languages.

\cite{poplack1980sometimes} worked on incorporating both linguistic and extra-linguistic factors into a single analytical model. This study concludes that there are no visibly ungrammatical combinations of the two languages and code-switching is independent of the bilingual ability of the speaker. \cite{sankoff1998formal} showed that there exists a constituent tree labeling, implying that around a switch point there is a constraint on an equivalence order in constituents. The above described linguistic theories are also used in \cite{di1986government} to identify governing relationships between constituents. \cite{belazi1994code} have demonstrated evidence that a constrained Universal Grammar needs refinement of f-selection in code-switching as compared to monolingual speech. \cite{sebba1998congruence} have proposed four categories for any switch point comprising of harmonization, neutralization, compromise, and blocking. 

\cite{gardner2004assumptions} have a rather interesting approach towards analyzing grammatical variants in code-switching based on pre-conceptualized assumptions. They claim that grammar in this context is subject to poly-idiolectal repertoires of bilingual speakers and sociolinguistic factors take precedence over grammatical factors. Hence they propose accounting for variability among the bilingual speakers. This same work was extended later to examine intra-sentential switching focusing on bilingual compound verbs and using grammatical knowledge.

The linguistic theories mentioned above were put to use in computational frameworks by \cite{bhat2016grammatical}. They address several issues such as the absence of literal level translation pairs, sensitivity to minor alignment errors and the under specification of the original models. The human evaluation of generated sentences reveals that the acceptability of code-switching patterns depend not only on socio-linguistic factors but also cognitive factors. This work was later extended in \cite{pratapa2018language} to perform language modeling by leveraging the theories discussed above to generate synthetic code-switched text. 

While on one hand, there are studies of formally constructing grammatical representations to understand the nature of code-switching, there is also work that focuses on understanding the psycho-linguistic aspect of this subject pertaining to how and when this occurs. 
There are studies pertaining to socially determined and pragmatic choices in the developmental perspective of switching in bilingual infants \cite{lanvers2001language}.

Another stream of work talks about the factors triggering code-switching that are attributed to `cognate' or trigger words including proper nouns, cognate content words with good and moderate form overlap, and cognate function words. \cite{backus2005codeswitching} have studied attested contact-induced changes based on prior linguistic theories regarding the types of structural changes in calques, distributions, frequencies, inventory and stability. \cite{soto2018role} explored three different hypothesis and presented empirical evidence for the same. They are the relationships between (i) cognate stimuli and code-switching, (ii) syntactic information and code-switching, (iii) entrainment in a code-switched conversation among bilinguals. The empirical evidence demonstrates that there is a strong correlation between precedence of cognates and code-switching, relationship between POS tags and code-switching and the convergence of the rate of entrianment in code-switching.

\cite{hartmann2018integrated} present an integrated representation of inter and intra-sentential phenomena as well as spoken and written modalities of code-switching. This is done in order to make better reuse of the minimally available code-switched data by analyzing various global dimensions such as modality, discourse, granularity, social familiarity and social hierarchy. These properties pave way to potentially footprint corpora and functional derivations. \cite{pratapa2017quantitative} present a systematic approach to analyze code-switching in conversations. Patterns of switching are analyzed in multi-party conversations from Hindi movie scripts to establish identity and social contexts. 

\subsection{Measuring the amount of code-switching}

Various metrics have been proposed to measure the amount of code-switching in corpora. The Code-mixing Index (CMI) \cite{gamback2016comparing} is an utterance and corpus level metric proposed to measure the amount of code-switching in corpora by using word frequencies. 
\cite{barnett2000lides} propose M-Index which quantifies the ratio of languages in the corpora based on the Gini coefficient to measure the inequality of the distribution of languages in the corpus. \cite{guzman2016simple} extend this metric to describe the probability of switching within a corpus by summing up the probabilities that there has been a language switch. This metric is termed Integration Index ( I-Index) and has values of in the range from 0 (a monolingual text in which no switching occurs) to 1.

\cite{guzman2017metrics} and \cite{guzman2017moving} also propose the following metrics: Language Entropy, Span Entropy, Burstiness and Memory. Language Entropy and Span Entropy are the number of bits needed
to represent the distribution of language spans. Burstiness quantifies whether the switching has
periodic character or occurs in bursts. Memory captures the tendency of consecutive language spans to be positively or negatively autocorrelated.

\cite{bullock2018predicting} propose techniques to automatically determine the matrix language of a code-switched utterance. Although the notion of the matrix language is based on the underlying grammar of the sentence, \cite{bullock2018predicting} show that the matrix language can be determined by word-count alone as an approximation. \cite{bullock2018should} characterize languages as being asymmetric or symmetric depending on whether code switching is insertional or alternating, and show that the same grammatical constraints hold in both cases.

\section{Data and resources}

Over the last few years, significant progress has been made in the fields of Speech Processing and Natural Language Processing mainly owing to the use of large and powerful Machine Learning models such as Deep Neural Networks (DNNs). DNNs typically require large labeled corpora for training, which can be found for a few languages such as US English, Mandarin and Modern Standard Arabic, which are commonly termed as high-resource languages. In the presence of large datasets, models can be trained to achieve high accuracies on tasks such as Automatic Speech Recognition, Machine Translation and Parsing.

However, most languages in the world do not have the necessary data and resources to create models with high enough accuracies to be used in real-world systems. The situation is even more stark for code-switched languages, since considerable care is taken to leave out foreign words while building monolingual resources. So, even if monolingual resources exist for one or more of the languages being mixed, code-switched speech and language resources are very scarce.

However, owing to the recent interest in code-switched speech and language processing, there are some speech and text data sets available for a few language pairs, which we describe next.

\subsection{Speech data}

Data used for building Automatic Speech Recognition (ASR) and Text to Speech (TTS) systems typically consists of recorded speech and the corresponding transcripts. For ASR systems, the speech may be spontaneous or read, and typically needs to be at least a few thousand hours to build systems that are usable. For TTS systems, a few hours of clean, well recorded speech from a single speaker is typically enough. Below is a list of code-switched data sets available for speech processing.

\begin{itemize}

\item SEAME \cite{lyu2010seame} is a corpus of Mandarin-English code-switching by bilinguals in Singapore and Malaysia, with Mandarin being the dominant language in the recordings. It consists of 63 hours of interviews and conversational speech from 97 speakers. \cite{lee2017review} describes the updated SEAME corpus with an additional 129 hours of speech.

\item The HKUST Mandarin-English Corpus \cite{li2012mandarin} also consists of interviews and conversational speech with 5 hours of transcribed and 15 hours of untranscribed speech.

\item The CECOS corpus \cite{shen2011cecos} contains 12 hours of prompted Mandarin-English speech from 77 speakers. 

\item The OC16-CE80 corpus is a 80 hour Mandarin-English corpus with English words embedded in Mandarin utterances \cite{wang2016oc16}.

\item The CUMIX Cantonese-English speech corpus \cite{chan2005development} contains 17 hours of code-switched speech read by 80 speakers. 

\item A Mandarin-Taiwanese corpus is described in \cite{lyu2006speech} containing 4000 utterances recorded by 16 speakers.

\item \cite{nakayama2018japanese} present an artificially generated Japanese-English code-switched corpus using a Japanese and English Text-to-speech system from a bilingual speaker. The corpus consists of 280k speech utterances.


\item BANGOR-MIAMI \cite{deuchar2014building} is a Spanish-English code-switched corpus consisting of 56 audio recordings and their corresponding transcripts. The recordings consist of informal conversations between two or more speakers, involving a total of 84 speakers.

\item A small Spanish-English corpus consisting of 40 minutes of spontaneous speech is described in \cite{franco2007baby}. 


\item \cite{ramanarayanan2017jee} describe an audio and video corpus of elicited code-switched Spanish-English and Hindi-English dialogues. The corpus consists of over 700 calls to an automated agent by workers on Amazon Mechanical Turk, although only a small subset of these have been transcribed.

\item The MSR Hindi-English database \cite{sivasankaran2018phone} consists of ~50 hours of conversational speech between Hindi-English bilinguals. There are around 500 speakers in the corpus. 


\item \cite{sreeram2018hindi} crawled blogs to collect Hindi-English code-switched utterances. 71 speakers recorded around 7000 utterances which were then transcribed and used to build an ASR system.

\item \cite{pandey2017adapting} describe the creation of a phonetically balanced Hindi-English corpus for code-switched ASR. This corpus contains read speech from 78 speakers, with each speaker having recorded around a minute of speech. The prompts have been collected from news websites and sampled for phonetic coverage.
 
\item \cite{dey2014hindi} collected a small Hindi-English corpus of student interviews. The corpus contains 3 minutes of transcribed speech from 9 speakers.

\item \cite{ahmed2012automatic} collected 1000 hours of Malay-English speech from 208 Chinese, Malay and Indian speakers.

\item An Egyptian Arabic-English speech corpus is described in \cite{hamed2018collection}. It consists of 5.3 hours of speech from interviews with 12 participants, of which 4.5 hours of speech has been transcribed.

\item The MCSM database
(Maghrebian Code-switching in Media) \cite{mohdeb2016arabic} consists of broadcasts from Morocco, Algeria and Tunisia with varying amounts of code-switching between French and Arabic. The FACST corpus (French Arabic Code-switching Triggered)\cite{amazouz2018french} consists of 7.3 hours of French-Algerian Arabic code-switched speech from 20 bilingual speakers.

\item \cite{ccetinouglu2017code} describes a corpus of Turkish-German conversational speech consisting of 5 hours of annotated speech data. The corpus is annotated with speech and orthography information, including inter and intra-word switch points.

\item \cite{yilmaz2017longitudinal} describe a corpus of radio broadcasts in Frisian covering a 50 year time span containing code-switching with Dutch. The corpus consists of 18.5 hours of speech annotated with speaker information, dialect and code-switching details and the presence of background noise/music.

\item A corpus of code-switched isiZulu-English consisting of transcribed speech from soap operas is described in \cite{van2016automatic}. It contains around 16 hours of transcribed speech with code-switching boundary annotations. This corpus is extended in \cite{van2018first, yilmaz2018building} to also include 14 hours of English-isiXhosa, English-Setswana, and English-Sesotho code-switched speech.

\item For Sepedi-English, two code-switching speech corpora are available \cite{modipa2013implications}. The Sepedi Radio corpus consists of broadcast speech which has been used for analyzing code-switching in this language pair. The Sepedi Prompted Speech Corpus consists of 10 hours of speech from 20 speakers. 
\item Microsoft and SpeechOcean.com recently released 60 hours of code-switched speech in three language pairs - Tamil-English, Telugu-English and Gujarati-English as part of a shared task on Spoken Language Identification \cite{shah2020first}.

\item Although no code-switched speech databases exist for Speech Synthesis, bilingual TTS databases are available from the same speakers in a number of Indian languages and English \cite{baby2016resources}.

\end{itemize}

\subsection{Text data}


In this section, we describe various resources that exist for processing code-switched text. Since the type of data and resources vary greatly with the task at hand, we describe them separately for each task.

\subsubsection{Language Identification (LID)}


Language ID data sets consist of code-switched sentences that are labeled at the word-level with language information. Conventional LID systems operate at the sentence or document level, which leads to the requirement of word-level LID for code-switched sentences. A couple of shared tasks played an important role in establishing datasets for language identification \cite{solorio2014overview} \cite{sequieraoverview}.

\begin{itemize}
  
    \item A predetermined set of 11 users of Facebook users were selected to search for publicly available content that resulted in 2335 posts and 9813 comments \cite{barman2014code}. Two levels of annotations were performed on this data comprising of different levels of code-switching and language tags including \textit{English,} \textit{Bengali,} \textit{Hindi,} \textit{Mixed,} \textit{Universal} and \textit{Undefined}. A similar approach is also followed by  \cite{das2014identifying} to collect more data from Bengali, Hindi and English with finer annotation schema catering to named entities and also explicitly annotating suffixes.
    
    \item \cite{patro2017all} used a two step process by crawling tweets related to 28 hashtags comprising of contexts ranging from sports, movies, religion, politics etc, that resulted in 811981 tweets. Identifying 3577 users from these tags, tweets from these users are crawled in order to gather more mixed language thus resulting in 725173 distinct tweets written in Roman script. These tweets are annotated with \textit{English}, \textit{Hindi} and \textit{Other} tags. 

    \item Another very large scale dataset that is is not explicitly targeted at code-switching but contains it is \cite{jurgens2017incorporating} that addresses curating socially representative text by taking into account geographic, social, topical and multilingual diversity. This corpus consists of Tweets from 197 countries in 53 languages.  
    
    \item In \cite{maharjan2015developing} authors employed different approaches to collect data in two different code mixed scenarios. For Nepali - English, they selected 42 twitter users and collected 2000 code switched tweets from each of them. For Spanish English they performed a geographical based search and used 50 tweets from 135 users after filtering based on manual inspection. 
\end{itemize}

\subsubsection{Named Entity Recognition (NER)}

Named Entity Recognition (NER) datasets for code-switching are similar to LID datasets, with word-level annotations.

\begin{itemize}
    \item A shared task was organized to address NER for code-switched texts using around 50k Spanish-English and around 10k Arabic-English annotated tweets \cite{aguilar2018named}.
    
    \item Twitter is a commonly used source of code-switched data. \cite{singh2018named} annotated 3,638 tweets with three Named Entity tags `Person', `Organization' and `Location' using the BIO scheme.
\end{itemize}

\subsubsection{Part of Speech (POS) Tagging}

POS tagging data sets consist of code-switched sentences tagged at the word level with POS information.

\begin{itemize}
    \item Public pages from Facebook pages of three celebrities and the BBC Hindi news page are used to gather 6,983 posts and comments and annotated with POS tags in addition to matrix language information \cite{vyas2014pos}.
    
    \item ICON 2015 conducted a shared task on POS tagging for which they released data in Hindi-English, Bengali-English, Tamil-English \cite{ghosh2016part}. The dataset contains 1k-3k annotated utterances for each language pair.
    
    \item Code-switched Turkish-German tweets were annotated based on Universal Dependencies POS tags and the authors proposed guidelines for the Turkish parts to adopt language-general heuristics to gather a corpus of 1029 tweets \cite{ccetinouglu2016part}. 
    
    
    \item 922 sentences of spoken Spanish-English conversational data is transcribed and annotated with  POS tags in \cite{solorio2008part}.
    
    \item \cite{jamatia2015part} gathered 1106 messages (552 Facebook posts and 554 tweets) in Hindi-English and annotated them with a Twitter specific tagset. \cite{jamatia2016collecting} describe an English-Bengali corpus consisting of Twitter messages and two English-Hindi corpora consisting of Twitter and Facebook messages tagged with coarse and fine grained POS tags. 
    
    \item \cite{soto2017crowdsourcing} crowd-sourced POS tags using the Universal POS tagset to annotate the BANGOR-MIAMI corpus which is a conversational speech dataset with Spanish-English code-switching.
    
    \item 
    \cite{diab2019creating} create a corpus of 886,252 tokens in Modern Standard Arabic-Egyptian Arabic annotated with POS as well as 16 code-switching tags. The code-switching tags include language ID information as well as special tags to label mixed words that contain morphemes from different languages. This dataset consists of tweets and sentences from news, commentaries and discussion forms. The annotation is being extended to other Arabic dialects such as Levantine, Iraqi, Gulf, Moroccan and Tunisian.

\end{itemize}

\subsubsection{Parsing}

Datasets for parsing contain code-switched sentences with dependency parses and chunking tags.

\begin{itemize}

    \item \cite{partanen2018dependency} have worked on using monolingual resources to parse low resource languages in the presence of code-switching. While the training data comprises of Russian UD v2.0 corpus with 3,850 sentences and 40 Komi sentences, the test set comprises of 80 Komi-Russian multilingual sentences and 25 Komi spoken sentences.
    
    \item \cite{bhat2017joining} have presented a dataset of 450 Hindi-English CM tweets for evaluation purposes annotated with dependency parse relations.
    
    \item A shallow parsing dataset comprising of 8450 tweets annotated with language id, normalized script, POS tagset and chunking tags is described in \cite{sharma2016shallow}.
    
    \item Code-switched test utterances for the NLmaps corpus are constructed by \cite{duong2017multilingual}. They use a parallel corpus of English and German utterances which share the same logical form to construct code-switched utterances. They use 1500 pairs of sentences from each language for training and 880 pairs for testing.
    
\end{itemize}

\subsubsection{Question Answering (QA)}

Question answering (QA) datasets typically consist of questions and answers that are in the form of articles, images, tuples etc. In case of code-switched QA, the questions are typically in code-switched form.

\begin{itemize}
    \item A first step towards creating a code-switched QA dataset was attempted by collecting 3000 questions from a version of a TV show \textit{``Who wants to be a Millionaire?''} and general knowledge questions from primary school textbooks for Hindi-English code-switching questions \cite{raghavianswer}. Out of the 3000 questions, 1000 unique questions are used in order to avoid any individual biases of language usage.
    
    \item In lieu of addressing the possibility of lexical bias from entrainment in \cite{raghavianswer}, another effort was made on a larger scale to collect 5933 questions for Hindi-English, Tamil-English, Telugu-English grounded on articles and images \cite{chandu2018code}. The final dataset included 1,694 Hindi-English, 2,848 Tamlil-English and 1,391 Telugu-English factoid questions and their answers \cite{chandu2019code}.
    
    \item Another section of efforts that move towards using monolingual data from English and weakly supervised and imperfect bilingual embeddings provided a test set of 250 Hindi-English code-switched questions mapped between SimpleQuestions dataset and Freebase tuples \cite{gupta2018transliteration}.
    
    \item One of the early efforts also include leveraging around 300 messages from social media platforms like Twitter and blogs to collect 506 questions from the domains of sports and tourism \cite{banerjeefirst}.
    
\end{itemize}

\subsection{Natural Language Inference}

\cite{khanuja2020new} present the first dataset for code-switched NLI, in which premises are taken from Bollywood (Hindi) movie scripts and annotators create hypotheses that entail or contradict the premises. The dataset contains 400 premises and around 2k hypotheses.

\subsection{Social media datasets}

Various datasets from social media such as Facebook and Twitter have been collected for different NLP tasks, which we describe in this section.

\cite{chakma2016cmir} collected 1959 Hindi-English tweets and asked annotators to rank tweets according to relevance for specific queries. 

\cite{bohra2018dataset} collect a Twitter corpus of around 4k tweets and annotate it for Hate Speech. \cite{vijay2018dataset} create a corpus of around 3k tweets for automated irony detection.


\section{Code-switched Speech and NLP Techniques}

\cite{thara2018code} present a brief survey of code-switching studies in NLP. \cite{ccetinouglu2016challenges} describe the challenges in computational processing of core NLP tasks as well as downstream applications. They highlight issues caused due to combining two languages at the lexical and syntactic level, using examples from several tasks and language pairs. 

In this paper, we provide a comprehensive description of work done in code-switched speech and NLP. Various approaches have been taken to build speech and NLP systems for code-switched languages depending on the availability of monolingual, bilingual and code-switched data. When there is a complete lack of code-switched data and resources, a few attempts have been made to build models using only monolingual resources from the two languages being mixed. 

Domain adaptation or transfer learning techniques can be used, wherein models are built on monolingual data and resources in the two languages and a small amount of `in-domain' code-switched data can be used to tune the models. 


Word embeddings have been used recently for a wide variety of NLP tasks. Code-switched embeddings can be created using code-switched corpora \cite{adithya2018}, however, in practice such resources are not available and other techniques such as synthesizing code-switched data for training such embeddings can be used. Massive multilingual models such as multilingual BERT \cite{devlin2018multilingual} have also been explored in code-switched NLP.

\subsection{Automatic Speech Recognition}

Since code-switching is a spoken language phenomenon, it is important that Automatic Speech Recognizers (ASRs) that are deployed in multilingual communities are able to handle code-switching. In addition, ASR systems tend to be the first step in a pipeline of different systems in applications such as conversational agents, so any errors made by ASR systems can propagate through the system and lead to failures in interactions.

Attempts have been made to approach the problem of code-switched ASR from the acoustic, language and pronunciation modeling perspectives.

Initial attempts at handling code-switched speech recognition identified the language being spoken by using a Language Identification (LID) system and then used the appropriate monolingual decoder for recognition. One  approach is to identify the language boundaries and subsequently use an monolingual ASR system to recognize monolingual fragments \cite{chan2004detection}. Another approach runs multiple recognizers in parallel with an LID system and uses scores from all the systems for decoding speech \cite{weiner2012integration}. In  \cite{ahmed2012automatic}, no LID system is used - instead, two recognizers in English and Malay are run in parallel and the hypotheses produced are re-scored to get the final code-switched recognition result. However, the disadvantages with multi-pass approaches are that errors made by the LID system are not possible to recover from. \cite{lyu2006speech} suggest a single-pass approach with soft decisions on LID and language boundary detection for Mandarin-Taiwanese ASR.

The choice of phone set is important in building ASR systems and for code-switched language pairs, the choice of phoneset is not always obvious, since one language can have an influence on the pronunciation of the other language. \cite{chan2009automatic} develop a cross-lingual phonetic Acoustic Model for Cantonese-English speech, with the phone set designed based on linguistic knowledge. \cite{yu2003chinese} present three approaches for Mandarin-English ASR - combining the two phone inventories, using IPA mappings to construct a bilingual phone set and clustering phones by using the Bhattacharyya distance and acoustic likelihood. The clustering approach outperforms the IPA-based mapping and is
comparable to the combination of the phone inventories. \cite{sivasankaran2018phone} describe approaches to combine phone sets, merge phones manually using knowledge and iterative merging using ASR errors on Hindi-English speech. Although the automatic approach is promising, manual merging using expert knowledge from a bilingual speaker performs best. \cite{vu2012first} use IPA, Bhattacharya distance and discriminative training to combine phone sets for Mandarin-English.  When code-switching occurs between closely related languages, the phone set of one language can be extended to cover the other, as is suggested in \cite{lyudovyk2014code} for Ukranian-Russian ASR. In this work, the Ukranian phone set and lexicon are extended to cover Russian words using phonetic knowledge about both languages.

\cite{modipa2013implications} describe an ASR system for Sepedi-English in which a single Sepedi lexicon is used for decoding. English pronunciations in terms of the Sepedi phone set are obtained by phone-decoding English words with the Sepedi ASR. \cite{pandey2017adapting} use a common Wx-based phone set for Hindi-English ASR built using a large amount of monolingual Hindi data with a small amount of code-switched Hindi-English data. \cite{yeh2015improved} use cross-lingual data sharing to tackle the problem of highly imbalanced Mandarin-English code-switching, where the speakers speak primarily in Mandarin.  In \cite{white2008investigation}, authors attempt to alleviate the problem of L2 word pronunciation by creating linguistically motivated pairwise mappings

When data from both languages is available but there is no or very little data in code-switched form, bilingual models can be built. \cite{vu2012first} train the Acoustic Model on bilingual data, while \cite{bhuvanagiri2010approach} and \cite{bhuvanagirir2012mixed} use existing monolingual models and with a phone-mapped lexicon and modified Language Model for Hindi-English ASR.
In \cite{taneja2019exploiting}, authors create synthetic code mixed speech by concatenating segments from different monolingual utterances and employ this to improve Hindi-English code mixed ASR performance. In \cite{acoustic_dataaugmentation_mandarinenglish_activelearning}, authors first detect real and untranscribed code mixed segments from online archives. They then employ semi supervised and active learning techniques to obtain transcriptions and use as augmented data to train code switched models. In \cite{yilmaz2018semi} authors follow semi supervised training and show that incorporating language and
speaker information is helpful while building bilingual acoustic models. In \cite{yilmaz2019multi} authors combine monolingual and bilingual graphs together with a unified acoustic model. 

\cite{winata2020meta} propose a technique known as meta transfer learning to select the best monolingual data for transfer that can improve code-switched models. \cite{zhang2020monolingual} describe the importance of data selection between subsets of English, Mandarin and code-switched datasets for improving Mandarin-English ASR and show that simply pooling all the data leads to worse results.

\cite{yilmaz2016investigating} build a bilingual DNN-based ASR system for Frisian-Dutch broadcast speech using both language-dependent and independent phones. The language dependent approach, where each phone is tagged with the language and modeled separately performs better. \cite{yilmaz2017exploiting} decode untranscribed data with this ASR system and add the decoded speech to ASR training data after rescoring using Language Models. In \cite{yilmaz2018acoustic}, this ASR is significantly improved with augmented textual and acoustic data by adding more monolingual data in Dutch, automatically transcribing untranscribed data, generating code-switched data using Recurrent LMs and machine translation.

\cite{yilmaz2018building,biswas2020semi} build a unified ASR system for five South African languages, by using interpolated language models from English-isiZulu,
English-isiXhosa, English-Setswana and English-Sesotho. This system is capable of recognizing code-switched speech in any of the five language combinations. 

\cite{guo2018study} use semi-supervised techniques to improve the lexicon, acoustic model and language model of English-Mandarin code-switched ASR. They modify the lexicon to deal with accents and treat utterances that the ASR system performs poorly on as unsupervised data. In \cite{nakayama2018speech} authors utilise ASR and TTS in a  semi supervised fashion to learn code switching. They further show that integrating language embeddings allows the framework to address even the language pairs not seen during training \cite{nakayama2019zero}.

In \cite{song2017multi} authors jointly train two Mandarin-English acoustic models that differ in the choice of acoustic units describing the salient acoustic and phonetic information. In \cite{song2017investigating}, they observe that sharing parameters between the primary and auxiliary tasks helps capture language switching information. 

Recent studies have explored end-to-end ASR for code-switching. Traditional end-to-end ASR models require a large amount of training data, which is difficult to find for code-switched speech. \cite{winata2018towards} propose a CTC-based model for Mandarin-English speech, in which the model is first trained using monolingual data and then fine-tuned on code-switched data. \cite{shaninvestigating} use transfer learning from monolingual models, wordpieces as opposed to graphemes and multitask learning with language identification as an additional task for Mandarin-English end-to-end ASR. In \cite{recognition_and_translation}, authors address the scenario where monolingual speakers attempt to comprehend code switched speech in the context of a dialog. To address this, they build a system to recognize code mixed speech and translate it to monolingual text. In \cite{constrained_output_embeddings}, authors present a hypothesis that the discrepancy between distributions of  token representations for different languages restricts end to end models. To alleviate this, they constrain the token representations using Shannon divergence and cosine distance. In \cite{codeswitched_asr_posterioradjustment}, authors perform a frame level language detection and adjust the posterior distribution with CTC conditioned on the language detection. \cite{zhang2020rnn} present an RNN-T model with language bias that can improve upon an RNN-T model without any LID information, without needing an explicit LID system. 

In \cite{luo2018towards}, authors explore using two types of units: characters for both Mandarin and English, characters for Mandarin and sub word units for English. In \cite{zeng2018end} authors employ BPE sub word units. In \cite{li2019towards}, authors employ a frame level language recognition system to seed CTC based acoustic model. 

Recent work by \cite{shah2020learning} showed that speech recognition models fine-tuned on code-switched data regress on monolingual speech. To alleviate this issue and build robust models that can improve on both monolingual and code-switched speech recognition, the authors propose using Learning Without Forgetting and adversarial training. \cite{reddy2020learning} extends this work by proposing a multi-task approach to domain adversarial training that shows further improvements on both monolingual and code-switched ASR.




As stated earlier, switching/mixing and borrowing are not always clearly distinguishable. Due to this, the transcription of code-switched and borrowed words is often not standardized, and can lead to the presence of words being cross-transcribed in both languages. \cite{brij2018} automatically identify and disambiguate homophones in code-switched data to improve recognition of code-switched Hindi-English speech.

\subsection{Language Modeling}

Language models (LMs) are used in a variety of Speech and NLP systems, most notably in ASR and Machine Translation. Although there is significantly more code-switched text data compared to speech data in the form of informal conversational data such as on Twitter, Facebook and Internet forums, robust language models typically require millions of sentences to build. Code-switched text data found on the Internet may not follow exactly the same patterns as code-switched speech. This makes building LMs for code-switched languages challenging. 

Monolingual data in the languages being mixed may be available, and some approaches use only monolingual data in the languages being mixed \cite{weng1997study} while others use large amounts of monolingual data with a small amount of code-switched data. 

Other approaches have used grammatical constraints imposed by theories of code-switching to constrain search paths in language models built using artificially generated data. \cite{li2012code} use inversion constraints to predict CS points and integrate this prediction into the ASR decoding process. \cite{li2014code} integrate Functional Head constraints (FHC) for code-switching into the Language Model for Mandarin-English speech recognition. This work uses parsing techniques to restrict the lattice paths during decoding of speech to those permissible under the FHC theory. \cite{li2013language} assign weights to parallel sentences to build a code-switched translation model that is used with a language model for decoding code-switched Mandarin-English speech.

\cite{ashutosh2017} show that a training curriculum where an Recurrent Neural Network (RNN) LM is trained ﬁrst with interleaved monolingual data in both languages followed by code-switched data gives the best results for English-Spanish LM. \cite{adithya2018} extend this work by using grammatical models of code-switching to generate artificial code-switched data and using a small amount of real code-switched data to sample from the artificially generated data to build Language Models. 

\cite{gebhardt2011speech} uses Factored Language Models for rescoring n-best lists during ASR decoding. The factors used include POS tags, code-switching point probability and LID. In \cite{adel2013recurrent}, \cite{adel2013combination} and \cite{adel2014features}, RNNLMs are combined with n-gram based models, or converted to backoff models, giving improvements in perplexity and mixed error rate. 

\cite{vu2014exploration, adel2015syntactic} investigate the importance of syntactic information such as Part-of-Speech(POS) in predicting the switching point. They observe that the switchig attitude is speaker dependent\cite{vu2013investigation}. 

\cite{van2017synthesising} synthesize isiZulu-English bigrams using word embeddings and use them to augment training data for LMs, which leads to a reduction in perplexity when tested on a corpus of soap opera speech.

In \cite{garg2018code} authors employ dual RNNs for language model training while  \cite{blaicher2011smt, winata2019code, chang2018code} investigate the applicability of artificially generated code mixed data for data augmentation. 

In \cite{chandu2018language}, authors show that encoding language information improves language model by learning code switch points. In \cite{gonen2018language} authors present a discriminative training based approach for model code mixed text. Alternatively, authors in \cite{zeng2017improving} propose to manipulate n gram based language model by employing clustering for the infrequent words. In \cite{winata2018code} authors present an approach using multi task learning by jointly learning language modeling as well POS tagging. 

\cite{lee2020modeling} use a bilingual attention language model that learns cross-lingual probabilities by using parallel data simultaneously along with the language modeling objective and achieves high reductions in perplexity over the SEAME corpus. 

\subsection{Code-switching detection from speech}

In \cite{piccinini2014prosodic} authors show that humans exploit prosodic cues to detect code mixing. They also show taht humans can anticipate switch poitns even in noisy speech.

As mentioned earlier, some ASR systems first try to detect the language being spoken and then use the appropriate model to decode speech. In case of intra-sentential switching, it may be useful to be able to detect the code-switching style of a particular utterance, and be able to adapt to that style through specialized language models or other adaptation techniques.

\cite{rallabandi2017building} look at the problem of language detection from code-switched speech and classify code-switched corpora by code-switching style and show that features extracted from acoustics alone can distinguish between different kinds of code-switching in a single language. 

In \cite{yilmaz2016code, yilmaz2018code} authors investigate the effectiveness of using retrained multilingual DNNS and augmenting the data for detecting the language. In \cite{lyu2010language,lyu2010analysis} authors employ word based lexical information 
\cite{mabokela2014modeling} build HMM based acoustic model followed by an SVM based decision classifier to identify the code mixing between Northern Sotho and English. 

\subsection{Speech Synthesis}

Most Text to Speech (TTS) systems assume that the input is in a single language and that it is  written in native script. However, due to the rise in globalization, phenomena such as code-switching are now seen in various types of text ranging from news articles through comments/posts on social media, leading to co-existence of multiple languages in the same sentence. Incidentally, these typically are the scenarios where TTS systems are widely deployed as speech interfaces and therefore these systems should be able to handle such input. Even though independent monolingual synthesizers today are of very high quality, they are not fully capable of effectively handling such mixed content that they encounter when deployed. These synthesizers in such cases speak out the wrong/accented version at best or completely leave the words from the other language out at worst. Considering that the words from other language(s) used  in such contexts are often the most important content in the message, these systems need to be able to handle this scenario better. 

Current approaches handling code-switching fall into three broad categories: phone mapping, multilingual or polyglot synthesis. In phone mapping, the phones of the foreign language are substituted with the closest sounding phones of the primary language, often resulting in strongly accented speech. In a multilingual setting, each text portion in a different language is synthesised by a corresponding monolingual TTS system. This typically means that the different languages will have different voices unless each of the voices is trained on the voice of same multilingual speaker. Even if we have access to bilingual databases, care needs to be taken to ensure that the recording conditions of the two databases are very similar. The polyglot solution refers to the case where a single system is trained using data from a multilingual speaker. Similar approaches to dealing with code-switching have been focused on assimilation at the linguistic level, and advocate applying a foreign linguistic model to a monolingual TTS system. The linguistic model might include text analysis and normalisation, a G2P module and a mapping between the phone set of the foreign language and the primary language of the TTS system \cite{tomokiyo2005foreign, campbell2001talking, badino2004language}. Other  approaches utilise cross-language voice conversion techniques \cite{mashimo2001evaluation} and adaptation on a combination of data from multiple languages\cite{latorre2005polyglot}. Assimilation at the linguistic level is fairly successful for phonetically similar languages \cite{badino2004language}, and the resulting foreign synthesized speech was found to be more intelligible compared to an unmodified non-native monolingual system but still retains a degree of accent of the primary language. This might in part be attributed to the  non-exact
correspondence between individual phone sets.

\cite{sitaram2015experiments} find from subjective experiments that listeners have a strong preference for cross-lingual systems with Hindi as the target language. However, in practice, this method results in a strong foreign accent while synthesizing the English words. \cite{elluru2013word,rallabandi2015iiit} propose a method to use a word to phone mapping instead, where an English word is statistically mapped to Indian language phones. 
 
\cite{thomas2018code} train speech synthesizers for Hindi-English, Tamil-English and Hindi-Tamil by randomizing the order of bilingual training data which are then used to synthesize monolingual and code-switched text. This leads to improvements in subjective metrics for the code-switched speech and marginal degradation in monolingual speech.
 
\cite{cao2019end} present an end-to-end code-switched TTS for Mandarin English, in which they use bilingual data with a shared encoder that contains language information and separate decoders. \cite{cao2020code} extend this approach to use a bilingual phonetic posteriorgram (PPG) to synthesize code-switched speech using only monolingual data. \cite{zhou2020multi} also use a language specific encoder along with a multi-head attention mechanism in the decoder resuling in large improvements in the SEAME corpus.

\subsection{Language Identification} 

The task of lexical level language identification (LID) is one of the skeletal tasks for the lexical level modeling of downstream NLP tasks. Most research has focused on word-level LID, although some work on utterance-level LID also exists. \cite{lignos2013toward} build tools for web-scale analysis of code-switching, using an utterance-level language identification system based on the language ratio of the two languages involved. A large amount of research in this area has been conducted due to shared tasks on word-level LID (\cite{solorio2014overview}, \cite{sequieraoverview}). 

Social media data, especially posts from Facebook was used to collect data for the task of LID \cite{barman2014code} of Bengali, Hindi and English code-switching. Techniques include dictionary based lookup, supervised techniques applied at word level along with ablation studies of contextual cues and CRF based sequence labeling approaches. Character level n-gram features and contextual information are found to be useful as features.

\cite{chittaranjan2014word} is among the first computational approaches towards determining intra-word switching by segmenting the words into smaller meaningful units through morphological segmentation and then performing language identification probabilistically. This was followed by intra-word approaches \cite{xia2016codeswitching, nguyen2016automatic}
and approaches that incorporate information beyond word level \cite{chanda2016unraveling,yeong2010language,elfardy2012token, yeong2011applying,yeong2014language, shirvani2016howard,dongen2017analysis,shrestha2016codeswitching,samih2016detecting,samih2016multilingual}. In addition to features, model based variants have been proposed by \cite{sikdar2016language,jaech2016neural,chang2014recurrent, jain2014language,chanda2016columbia}.

\cite{jhamtani2014word} make use of patterns in language usage of Hinglish along with the consecutive POS tags for LID. \cite{das2014identifying} have also experimented with n-gram modeling with pruning and SVM based models with feature ablations Hindi-English and Bengali-English LID.  \cite{patro2017all} have worked on re-defining and re-annotating language tags from social media cues based on cultural, core and therapeutic borrowings. \cite{jurgens2017incorporating} have introduced a socially equitable LID system known as EQUILID by explicitly modeling switching with character level sequence to sequence models to encompass dialectal variability in addition to code-switching. \cite{king2013labeling} present a weakly supervised approach with a CRF based on a generalization expectation criteria that outperformed HMM, Maximum Entropy and Naive Bayes methods by considering this a sequence labeling task.

Recently, POS tagging has also been examined as a means to perform language identification in code-switched scenarios \cite{ansari2020analysis}. To this end, they have collected a Devanagari corpus and annotated it with POS tags followed by  transliterating it into Roman text. The complementary English data is annotated with POS tags as well. Several classical approaches including SVM, Decision Trees, Logistic Regression and Random Forests have been experimented with. The feature set that included POS tags along with the word length and the word itself with a random forest resulted in the highest performance. Hence monolingual data with corresponding POS tags seem useful in performing language identification of code-switched text.


\subsection{Named Entity Recognition}

Another sequence labeling task of interest is Named Entity Recognition (NER). \cite{aguilar2018named} organized a shared task on NER in code-switching by collecting data from tweets for Spanish-English and Arabic-English. \cite{attia2018ghht} augmented state-of-the-art character level Convolutional Neural Networks (CNNs) with Bi-LSTMs followed by a CRF layer, by enriching resources from external sources by stacking layers of pre-trained embeddings, Brown clusters and gazetteer lists. \cite{geetha2018tackling} attempted to build models from observations from data comprising of less than 3\% of surface level Named Entities and a high Out of Vocabulary (OOV) percentage. To address these issues they rely on character based  BiLSTM models and leveraging external resources. Prior to this shared task, \cite{aguilar2017multi} posed this task as a multi-task learning problem by using a character level CNNs to model non-standard spelling variations followed by a word level Bi-LSTM to model sequences. This work also highlights the importance of gazetteer lists since it is similar to a low resource setting. 

\cite{zirikly2015named} studied Arabic text on social media by exploring the influence of word embedding based representations on NER. Along similar lines, \cite{zea2016spanish} also investigated how word representations are capable of boosting semi-supervised approaches to NER. \cite{singh2018named} collected tweets from topics like politics, social events, sports and annotated them with three Named Entity Tags in the BIO scheme and explored CRF, LSTM and Decision Tree methods. Formal and informal language specific features were leveraged to employ Conditional Random Fields, Margin Infused Relaxed Algorithm, Support Vector Machines and Maximum Entropy Markov Models to perform NER on informal text in Twitter \cite{etter2013nerit}. \cite{sabtytechniques} collected a dataset in an attempt to benchmark for the task of Named Entity Recognition in Arabish from three different sources: Twitter, transcribed conversational speech and translating a standard NER dataset. The dataset comprises of 6k sentences with 130k tokens. The baseline model itself is a BiLSTM-CRF which is one of the heavily investigated architectures in the task of NER. On top of this, they have adopted the FLAIR framework \cite{akbik2018coling} to investigate different types of embeddings along with pooled datasets. They have also experimented with word embeddings that are not only traditionally used and also more recently used such as contextual embeddings in their architecture and discovered that a combination of both performed better. \cite{winata2018bilingual} extend the LSTM architecture to combat high percentage of out of vocabulary words in code-switched data using transfer learning with bilingual character representations. Additionally, they also remove the noise with normalization of the spellings. Alternative to the fusion approach see above, \cite{wang2018code} utilize the self attention mechanism over the charcacter based embeddings. The final embedding representation is obtained by feeding these word and character based embeddings through a stacked BiLSTM with residual connections. Inspired by this, \cite{winata2019learning} proposed multilingual meta embeddings that extend the scope to other related and similar languages. They circumvent the problem of lexical level language identification using the same self attention mechanism on pre-trained word embeddings. \cite{winata2019hierarchical} propose the use of hierarchical meta-embeddings that combine word and sub-word level embeddings to achieve SOTA performance on English-Spanish NER.

\subsection{POS Tagging} 

Recently there has been interest in code-switched structured prediction tasks like POS tagging and parsing. \cite{vyas2014pos} used a dual mechanism of utilizing both a CRF++ based tagger and a Twitter POS tagger in order to tag sequences of mixed language. The same work also proposed a dataset that is obtained from Facebook that is annotated at a multi-level for the tasks of LID, text normalization, back transliteration and POS tagging. They claim that joint modeling of all these tasks is expected to yield better results. \cite{ccetinouglu2016part} presented POS annotation for Turkish-German tweets that align with existing language identification based on POS tags from Universal Dependencies. \cite{solorio2008part} explored the exploitation of monolingual resources such as taggers (for Spanish and English data) and heuristic based approaches in conjunction with machine learning techniques such as SVM, Logit Boost, Naive Bayes and J48. This work shows that many errors occur in the presence of intra-sentential switching thus establishing the complexity of the task.

\cite{jamatia2015part} have also gathered data from social media platforms such as Facebook and Twitter and have annotated them at coarse and fine grained levels. They focus on comparing language specific taggers with ML based approaches including CRFs, Sequential Minimal Optimization, Naive Bayes and Random Forests and observ that Random Forests performed the best, although only marginally better than combinations of individual language taggers. \cite{soto2017crowdsourcing} use crowd-sourcing for annotating universal POS labels for Spanish-English speech data by splitting the task into three subtasks. These are 1. labeling a subset of tokens automatically 2. disambiguating a subset of high frequency words 3. crowd-sourcing tags by decisions based on questions in the form of a decision tree structure. The choice of mode of tagging is based on a curated list of words.

\cite{barman2016part} use a stacked model technique and compare them to joint modeling and pipeline based techniques to find that that best stacked model that utilizes all features outperform the joint and pipeline-based models.

\cite{van2020lexical} carry out normalization of code-switched data and assess the impact on POS tagging as a downstream task. They find that automatic normalization leads to a performance gain in POS tagging.

\subsection{Parsing}

\cite{goyal2003bilingual} worked on bilingual syntactic parsing techniques for Hindi-English code-switching  using head-driven phrase structure grammar. The parses in cases of ambiguities are ordered based on ontological derivations from WordNet through a Word Sense Disambiguator \cite{sharma2002saarthaka}. However, there is an assumed external constraint in this work, where the head of the phrase determines the syntactic properties of the subcategorized elements irrespective of the languages to which these words belong.

\cite{bhat2017joining} leveraged a non-linear neural approach for the task of predicting the transitions for the parser configurations of arc-eager transitions by leveraging only monolingual annotated data by including lexical features from pre-trained word representations. \cite{sharma2016shallow} also worked on a pipeline and annotating data for shallow parsing by labeling three individual sequence labeling tasks based on labels, boundaries and combination tasks where a CRF is trained for each of these tasks.

\cite{duong2017multilingual} performed multilingual semantic parsing using a transfer learning approach for code-switched text utilizing cross lingual word embeddings in a sequence to sequence framework. \cite{partanen2018dependency} compared different systems for dependency parsing and concluded that the Multilingual BIST parser is able to parse code-switched data relatively well. 

\cite{bhat2018universal} present a Universal Dependencies dataset in Hindi-English and a neural stacking model for parsing with a new decoding scheme that outperforms prior approaches.

\subsection{Question Answering}

So far, we have seen individual speech and NLP applications which can be used as part of other downstream applications. One very impactful downstream application of casual and free mixing beyond mere borrowing in terms of information need is Question Answering (QA). This is especially important in the domains of health and technology where there is a rapid change in vocabulary thereby resulting in rapid variations of usage with mixed languages. One of the initial efforts in eliciting code-mixed data to perform question classification was undertaken by \cite{raghavianswer}. This work leveraged monolingual English questions from websites for school level science and maths, and from Indian version of the show \textit{`Who wants to be a Millionaire?'}. Crowd-workers are asked to translate these questions into mixed language in terms of how they would frame this question to a friend next to them.

Lexical level language identification, transliteration, translation and adjacency features are used to build an SVM based Question Classification model for data annotated based on coarse grained ontology proposed by \cite{li2002learning}. Since this mode of data collection has the advantage of gathering parallel corpus of English questions with their corresponding code-switched questions, there is a possibility of lexical bias due to entrainment. In order to combat this, \cite{chandu2018code} discussed techniques to crowd-source code-mixed questions based on a couple of sources comprising of code-mixed blog articles and based on certain fulcrum images. They organized the first edition of the code-mixed question answering challenge where the participants used techniques based on Deep Semantic Similarity model for retrieval and pre-trained DrQA model fine-tuned on the training dataset. An end-to-end web based QA system WebShodh is built and hosted by \cite{chandu2017webshodh} which also has an additional advantage of collecting more data. 

\cite{gupta2018transliteration} trained TripletSiamese-Hybrid CNN to re-rank candidate answers that are trained on the SimpleQuestions dataset in monolingual English as well as with loosely translated code-mixed questions in English thereby eliminating the need to actually perform full fledged translation to answer queries. \cite{banerjeefirst} gathered a QA dataset from Facebook messages for Bengali-English CM domain. In addition to this line of work, there were efforts for developing a cross-lingual QA system where questions are asked in one language (English) and the answer is provided in English but the candidate answers are searched in Hindi newspapers \cite{sekine2003hindi}.

\cite{pingali2008dictionary} presented a query oriented multi-document summarization system for Telugu-English with a dictionary based approach for cross language query expansion using bilingual lexical resources. Cross language QA systems are explored in European languages as well \cite{neumanncross}, \cite{magninimultiple}.

\subsection{Sentiment Analysis/stance detection}

\cite{vilares2016cs} provide a benchmarking dataset to perform sentiment analysis on 3,062 English-Spanish tweets. The annotations are based on SentiStrength into the labels of positive, negative and neutral classes. The same group later extended this work to compare code-switching in monolingual and multilingual settings \cite{vilares2015sentiment, vilares2017supervised}. The comparisons made between a multilingual model trained on a multilingual dataset, separate monolingual models, and a monolingual model that is triggered based on the language identification demonstrate the effectiveness of the multilingual model to deal with code-switched scenarios. \cite{padmaja2020sentiment} use a more classical word probabilities based approach to determine the sentiment of a tweet about a movie. In specific, this is performed for tweets in Telugu-English mixed data by transliterating each Roman word to the corresponding Telugu script and computing the probability of the word in each class. \cite{patra2018sentiment} conducted a shared task for sentiment analysis of social media data in two language pairs. The dataset used for the shared task includes around 12k and 2500 tweets released for training in Hindi-English and Bengali-English respectively. The best performing system of the shared task used word and character level n-gram features with an SVM classifier. A similar trend is observed by \cite{mandal2018analyzing} while comparing the models of Naive Bayes and SVM to perform sentiment classification on movie reviews in Bengali-English. 

Contrary to this, \cite{lal2019mixing} use CNNs to model sub-word level representations. These are then given to a dual encoder, which both capture the sentiment at the sentence level and at the sub-word level. Similarly, \cite{utsav2020stance} approached this using multitask learning over a CNN based encoder to classify the stance taken on a popular issue of `Demonetization'. The auxiliary task is posed as a manipulation of the primary task by combining the labels. 

Extending this to emotion detection, there have been several attempts to model this as a graph problem. \cite{lee2015emotion} present a proposition of scheme to annotate data collected with emotions for Chinese-English corpus specifically. The schema is developed to address the choice of text in which a sentiment is expressed. This can be either in one of Chinese, or English text, or using both the languages, or using a mixed language text. \cite{wang2015emotion} gathered a dataset from \textit{Weibo.com} which have labelled emotions which are then self aligned among the languages using statistical machine translation paradigm. They use label propagation over a bipartite graph constructed on bilingual and sentiment information. They have extended this work to joint factor graph model \cite{wang2016emotion} between the two kinds of information identifying the necessity of correlating different emotions as well in addition to sentiment and languages. Along very similar lines, \cite{supraja2017emotion} use belief propagation over the factor graphs which poses this as a dynamic programming approach to query a graphical model. The graph itself is constructed as joint factor graph model by utilizing both the bilingual word information and the emotion related information. 

\subsection{Hate Speech Detection}

In \cite{rajput2020transfer}, authors employ transfer learning. They first train a CNN based model on a large corpus of hateful tweets as source task followed by fine tuning on a transliterated set in the same language. In \cite{khandelwal2020unified}, authors use a combination of psycho-linguistic feature and basic features and perform model averaging. In \cite{santosh2019hate}, authors investigate both  hierarchical employing phonemic units and sub world level models to detect hate speech from code mixed data.




\subsection{Natural Language Inference}

\cite{khanuja2020new} present the first work on code-switched NLI, where the task is to predict if a hypotheses entails or contradicts the given premise, which is in the form of a conversation taken from Bollywood (Hindi) movies. They fine-tune multilingual BERT for the task, however, the accuracy of this model is only slightly better than chance showing that NLI is a very challenging problem for code-switched NLP.

\subsection{Machine Translation} 

\cite{sinha2005machine} developed a machine translation scheme for translating Hinglish into pure English and pure Hindi forms by performing cross morphological analysis. \cite{johnson2017google} show that a zero shot Neural Machine Translation system can also deal with code-switched inputs, however, the results are not as good as monolingual inputs.

\subsection{Dialogue and discourse}

\cite{rao2018study} study lexical and prosodic features of code-switched Hindi-English dialogue and find that the embedded language (English) fragments are spoken more slowly and with more vocal effort, and pitch variation is higher in the code-switched portion of the dialogues compared to the monolingual parts.

\cite{bawa2018accommodation} treat code-choice as linguistic style and study accommodation across turns in dialogues in Spanish-English and Hindi-English. They find that accommodation is affected by the markedness of the languages in context and is sometimes seen after a few turns, leading to delayed accommodation. 

In \cite{ahn2020code}, authors investigate the effectiveness of linguistically motivated strategies of code mixing in a goal oriented dialogue setting. 

Cross-lingual Question Answering systems were extended to dialog systems for railway inquiries \cite{reddy2006dialogue}. Recently, there has been an attempt to create code-mixed version of goal oriented conversations \cite{banerjee2018dataset} from the DSTC2 restaurant reservation dataset.

As we have discussed earlier, code-switching is a phenomenon observed in informal scenarios, which implies that it is a suitable setting in conversational speech. In coherence to this thought, there has been work on incorporating and comparing text and speech based features in identifying language at turn level in a dialog \cite{ramanarayanan2019automatic}. Hence this work focuses on inter-sentential switching as opposed to intra-sentential switching. This work demonstrated the efficacy of i-Vector features in comparison to spectral features for speech segments. While the best performing system is based on text features, this work equips the field to work with speech based features when transcriptions are not available. 

\subsection{User Interfaces}

While languages that are being mixed may share the same script (such as in the case of English and Spanish), this is not true for many language pairs that are frequently code-switched, particularly when the languages are not related to each other. In such cases, users may choose to use the same script to write both languages, or use a mixed script. This has implications not only in how to process mixed languages, but also on how to display them. \cite{srivastava2020understanding} presents a study on the interaction between script-mixing and language mixing for Hindi-English and shows that script choice may be used for emphasis, disambiguation and marking whether a word is borrowed or not. 

\subsection{Optical Character Recognition}

\cite{garrette2015unsupervised} extend a standard OCR model to enable transcription of code-switched text by jointly performing transcription with word-level language identification. The model provides significant error reductions in historical texts.

\subsection{Improving cross-lingual models}

Recently, code-switched text has been used to improve the performance of cross-lingual systems. Code-switching is seen as a bridge to anchor representations in different languages so that they can come closer in a common space and lead to improved performance in cross-lingual NLP tasks \cite{qin2020cosda}. \cite{khudabukhsh2020harnessing} use code-switched text along with an English language identifier to retrieve documents written in Romanized Hindi. \cite{yang2020alternating} use alternating language modeling by artificially generating code-switched text using phrase alignments between parallel sentences to improve performance on cross-lingual tasks such as XNLI \cite{conneau2018xnli}. \cite{qin2020cosda} use a similar approach in which they synthesize random code-switched sentences in multiple languages to improve zero-shot performance on XNLI. 

\section{Evaluation of Code-switched Systems}

Much of the progress in a new field can be shaped by shared tasks in which common datasets are released and participants compete to build systems for a specific task. There have been several shared tasks conducted for code-switched text processing, and a few shared tasks for code-switched speech processing over the last few years. Shared tasks for code-switched NLP have included Language Identification \cite{diab2014proceedings, solorio2014overview, molina2019overview}, transliterated search \cite{roy2013overview}, code-mixed entity extraction \cite{rao2016cmee}, mixed script information retrieval \cite{sequiera2015overview, banerjee2016overview}, POS tagging \cite{jamatia2016task}, Named Entity Recognition \cite{aguilar2018named}, Sentiment Analysis \cite{patra2018sentiment} and Question Answering \cite{chandu2018code, chandu2019code}. There have been fewer shared task for code-switched speech processing, however, the Blizzard challenge 2014 had a code-switched speech synthesis task \cite{prahallad2014blizzard}, and code-switched ASR challenges have been conducted for Mandarin English \cite{wang2016oc16, shi2020asru}. Recently, a spoken Language Identification challenge was conducted for inter and intra-utterance LID \cite{shah2020first} in three code-switched language pairs.

Each of these shared task have spurred research in their respective sub-areas of code-switched speech and NLP. However, it is not clear how well these individual models can generalize across different tasks and language pairs. To address this gap, benchmarks for evaluating code-switching across different NLP have been proposed.

The GLUECoS benchmark \cite{khanuja2020gluecos} consists of 11 datasets spanning different tasks for code-switching across two language pairs Spanish-English and Hindi-English, including a new task for code-switching, Natural Language Inference (NLI). The GLUECoS benchmark aims to add more tasks to evaluate the general language understanding capabilities of models, including tasks such as Question Answering, Natural Language Generation and Summarization, Machine Translation and NLI. The LINCE benchmark \cite{aguilar2020lince} consists of 10 datasets across 5 language pairs. The tasks include LID, NER, POS tagging and Sentiment Analysis.

Evaluations conducted on the benchmarks described above indicate that massively multilingual contextual language models such as multilingual BERT \cite{devlin2018multilingual} outperform cross-lingual models and other task-specific models. These models can be further improved by adding synthetic code-switched data to pre-training, as shown in \cite{khanuja2020gluecos}. While models on some word level tasks such as Language Identification and Named Entity Recognition reach high accuracy, the performance on harder tasks like Sentiment Analysis, Question Answering and NLI is much worse and there is a large gap between the performance of models on monolingual tasks compared to code-switched tasks. This indicates that massive multilingual models do not perform as well on code-switching as they do on monolingual or even cross-lingual tasks. However, pre-training or fine-tuning such models on synthetic code-switched data in the absence of real code-switched data seems to a be promising future direction.

\section{Challenges and Future Directions}


Although code-switching is a persistent phenomenon through out the whole world, access to data will always be limited.  Monolingual corpora will always be easier to find as monolingual discourse is more common in formal environments and hence more likely to be archived. Code-switching data, by its nature of being used in more informal contexts, is less likely to be archived and hence harder to find as training data. As code-switching is more likely to be used in less task specific contexts, with less explicit function it may also be more difficult to label such data.

Most code-switching studies focus on pairs with one high resource language (e.g. English, Spanish, MSA, Putonghua) and a lower resource language, realistically the position is much more complex than that.  Although we consider Hinglish data low resourced, there are many other Northern Indian languages that are code-switched with Hindi and access to that data is even harder.
Thus code-switching studies will inherently always be data starved and our models must therefore expect to work with limited data.

Most current work in code-switching looks at one particular language pair.  It is not yet the case that architectures for multiple pairs are emerging, except perhaps within the Indian sub-continent where there are similar usage patterns with English and various regional languages.  However it is clear that not all code-switching is the same.  Relative fluency, social prestige, topical restrictions and grammatical constraints have quite different effects on code-switching practices thus it is hard to consider general code-switching models over multiple language pairs. 

We should also take into the account that as code-switching is more typical in less formal occasions, there are some tasks that are more likely to involve code-switching that others.  Thus we are unlikely to encounter programming languages that use code-switching, but we are much more likely to encounter code-switching in sentiment analysis.  Likewise analysis of parliamentary transcripts are more likely to be monolingual, while code-switching is much more likely in social media.  Of course its not just the forum that affects the distribution, the topic too may be a factor.

These factors of use of code-switch should influence how we consider development of code-switched models. Although it may be possible to build end-to-end systems where large amounts of code-switching data is available, in well-defined task environments, such models will not have the generalizations we need to cover the whole space. For models that can make use of large amounts of unlabeled data for training, generating synthetic code-switched data may be a promising direction. However, most models of code-switched data generation rely on syntactic constraints and do not take into account sociolinguistic factors that affect code-switched language. Building models that are capable of incorporating these factors could lead to more realistic data generation, which could lead to better models for code-switched speech and NLP.

It is not yet clear yet from the NLP point of view if code-switching analysis should be treated primarily as a translation problem, or be treated as a new language itself.  It is however likely as with many techniques in low-resource language processing, exploiting resources from nearby languages will have an advantage.  It is common (though not always) that one language involved in code-switching has significant resources (e.g. English, Putonghua, Modern Standard Arabic). Thus transfer learning approaches are likely to offer short term advantages. Also given the advancement of language technologies, particularly due to the rise of massively multilingual models, developing techniques that can work over multiple pairs of code-switched languages may lead to faster development and generalization of the field.

Evaluating code-switched speech and NLP is challenging due to the lack of standardized datasets. Although initial attempts at creating benchmarks have been made, a comprehensive evaluation of code-switched systems across speech and NLP tasks in many typologically different language pairs is required. Such evaluation benchmarks are even more important due to the prevalence of multilingual models that perform zero-shot cross-lingual transfer well, and are also expected to perform well on code-switched languages.

Speech and language technologies for code-switching is not yet a mature field.  It is noted that the references to work in this article are for the most part the beginnings of analysis. They are investigating the raw tools that are necessary in order for the development of full systems.  While there has been a lot of work on individual Speech and NLP systems for code-switching, there are no end-to-end systems that can interact in code-switched language with multilingual humans. Specifically we are not yet seeing full end-to-end digital assistants for code-switched interaction, or sentiment analysis for code-switched reviews, or grammar and spelling for code-switched text. This is partly due to lack of data for such end-to-end systems, however, a code-switching intelligent agent has to be more than just the sum of parts that can handle code-switching. To build effective systems that can code-switch, we will also have to leverage the work done in sociolinguistics to understand how, when and why to code-switch.

\section*{References}

\bibliography{mybibfile}

\end{document}